\documentclass[conference]{IEEEtran}
\IEEEoverridecommandlockouts

\usepackage{cite}
\usepackage{amsmath,amssymb,amsfonts}
\usepackage{algorithmic}
\usepackage{graphicx}
\usepackage{textcomp}
\usepackage{xcolor}
\usepackage{cite}
\usepackage{amsmath,amssymb,amsfonts,amsthm}

\usepackage{xcolor}
\usepackage{hyperref}

\hypersetup{
    colorlinks=true,
    urlcolor=blue
}

\usepackage{algorithmic}
\usepackage{graphicx}
\usepackage{textcomp}
\usepackage{xcolor}
\usepackage{pgfplots}
\usepackage{float}
\usepackage{booktabs}
\usepackage{subcaption}
\usepackage{float}
\def\BibTeX{{\rm B\kern-.05em{\sc i\kern-.025em b}\kern-.08em
    T\kern-.1667em\lower.7ex\hbox{E}\kern-.125emX}}

\begin{document}

\title{Large Language Models for Math Education in Low-Resource Languages: A Study in Sinhala and Tamil}

\author{
\IEEEauthorblockN{
\begin{tabular}{c c}
\begin{tabular}{c}
1\textsuperscript{st} Sukumar Kishanthan \\
\textit{Department of Computer Science and Engineering} \\
\textit{University of Moratuwa} \\
Moratuwa, Sri Lanka \\
sukumar@cse.mrt.ac.lk
\end{tabular}
&
\begin{tabular}{c}
2\textsuperscript{nd} Kumar Thushalika \\
\textit{Department of Electrical and Information Engineering} \\
\textit{University of Ruhuna} \\
Galle, Sri Lanka \\
thushalika\_k\_e23@engug.ruh.ac.lk
\end{tabular}
\\
\\
\begin{tabular}{c}
3\textsuperscript{rd} Buddhi Jayasekara \\
\textit{Faculty of Information Technology and Communication} \\
\textit{Sciences} \\
\textit{Tampere University} \\
Tampere, Finland \\
buddhi.jayasekara@tuni.fi
\end{tabular}
&
\begin{tabular}{c}
4\textsuperscript{th} Asela Hevapathige \\
\textit{School of Computing} \\
\textit{Australian National University} \\
Canberra, Australia \\
asela.hevapathige@anu.edu.au
\end{tabular}
\end{tabular}
}
}
\maketitle

\begin{abstract}
Large language models (LLMs) have achieved strong results in mathematical reasoning, and are increasingly deployed as tutoring and learning support tools in educational settings. However, their reliability for students working in non-English languages, especially low-resource languages, remains poorly understood. We examine this gap by evaluating mathematical reasoning in Sinhala and Tamil---two languages widely used in South Asian schools but underrepresented in artificial intelligence (AI) research. Using a taxonomy of six math problem types, from basic arithmetic to complex unit conflict and optimization problems, we evaluate four prominent large language models. To avoid translation artifacts that confound language ability with translation quality, we construct a parallel dataset in which each problem is independently authored in Sinhala and Tamil by native speakers, and in English by fluent speakers, all with strong mathematical backgrounds. Our analysis demonstrates that while basic arithmetic reasoning transfers robustly across languages, complex reasoning tasks show significant degradation in Tamil and Sinhala. The pattern of failures varies by model and problem type, suggesting that strong performance in English does not guarantee reliable performance across languages. These findings have direct implications for the deployment of AI tools in multilingual classrooms, and highlight the need for language-specific evaluation before adopting large language models as math tutoring aids in non-English educational contexts.
\end{abstract}


\begin{figure*}[t]
\centering
\includegraphics[width=\textwidth]{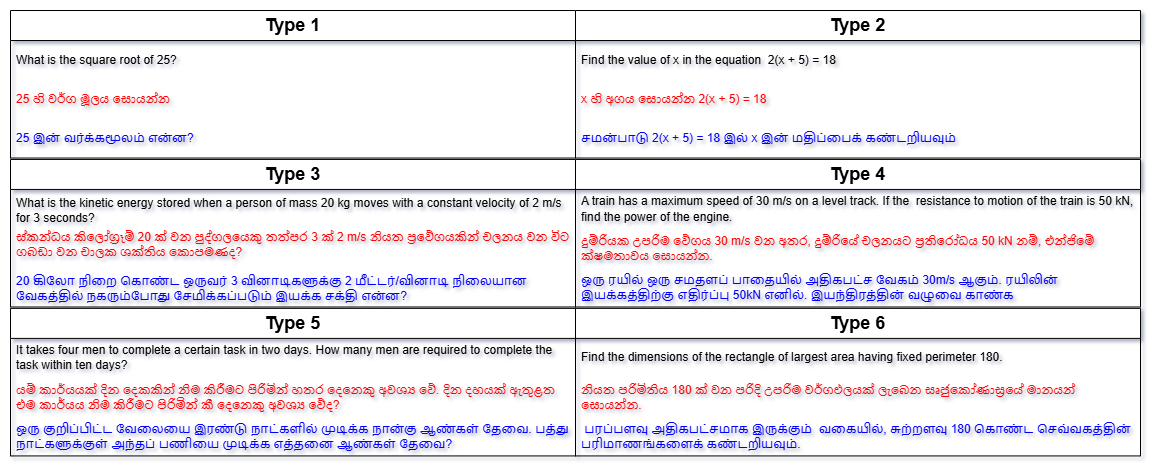}
\caption{Sample problems from each of the six problem types in \textcolor{black}{English}, \textcolor{red}{Sinhala}, and \textcolor{blue}{Tamil}, illustrating the taxonomy's coverage from basic arithmetic (Type 1) to complex optimization (Type 6).}
\label{fig:questions}
\end{figure*}

\section{Introduction}
\label{sec:intro}

Mathematical word problem solving has proven to be a challenging task for natural language processing techniques, requiring models to parse linguistic descriptions and perform complex multi-step reasoning \cite{dewappriya2018unit,hevapathige2018two,saraf2024towards}. Large language models (LLMs) have achieved impressive results on English mathematical reasoning benchmarks~\cite{yan2025survey,liu2025mathematical,wang2025survey}, and are now widely deployed as tutoring systems, homework assistants, and automated graders in educational settings \cite{ahn2024large,stamenkova2025large,lee2025can,tan2025elf}. But how well do these abilities extend to low-resource languages like Sinhala and Tamil remains underexplored. Compared to English, these South Asian languages present certain challenges for LLMs. They have training data scarcity, complex morphology, extensive case marking, and flexible word order that can separate numbers from operations \cite{hock2016languages,arora2022computational}. For instance, Tamil uses a different numeral system alongside Arabic numerals, while Sinhala's rich inflection can express quantitative relationships in unique ways. Additionally, these languages often structure mathematical relationships differently; a simple prepositional phrase in English might require complex constructions in Tamil or case-marked noun phrases in Sinhala. This raises an important question for education: \emph{Can students in Tamil- or Sinhala-medium schools rely on these tools for math support, or does performance drop in ways that undermine their usefulness in the classroom?}

Yet answering this question is harder than it appears. Most existing evaluations test models in Sinhala and Tamil by simply translating English math word benchmarks such as GSM8K \cite{cobbe2021training} or MGSM~\cite{shi2023language}, thereby introducing artifacts such as unnatural phrasing, inconsistent number formats, and culturally inappropriate contexts that confound language ability with translation quality. Beyond this, these evaluations usually report only overall accuracy scores, treating mathematical reasoning as a single capability. This approach misses important details: a model might excel at basic arithmetic in a given language while struggling with complex problems. This paper addresses both issues through three contributions:

\begin{enumerate}
    \item We develop a taxonomy of six math word problem types, each targeting a specific mathematical skill and defined by formal structural properties.
    
    \item We create a parallel dataset with problems natively written in English, Sinhala, and Tamil by fluent speakers with mathematical training, avoiding translation artifacts.
    
    \item We test four leading LLMs using zero-shot prompting, analyzing performance across both problem types and languages to identify which reasoning skills are most vulnerable to cross-lingual degradation.
\end{enumerate}

Our results show that cross-lingual performance loss varies systematically across different types of mathematical reasoning. Some problem categories transfer robustly across languages, while others show substantial drops that differ between model architectures. These patterns remain hidden under aggregate metrics, demonstrating the need for more detailed approaches to evaluate multilingual mathematical reasoning, and for more careful consideration of which tasks can safely be delegated to AI tools in non-English classroom settings.

The remainder of the paper is organised as follows. Section~\ref{sec:related} reviews related work. Section~\ref{sec:methodology} describes the problem taxonomy, dataset construction, and evaluation protocol. Section~\ref{sec:results} presents results and analysis, and Section~\ref{sec:conclusion} discusses conclusions, limitations and future directions for our work.

\section{Related Work}\label{sec:related}

\subsection{Mathematical Reasoning in LLMs}

LLMs have been evaluated and have shown impressive performance in solving math word problems \cite{yao2023solving,anantheswarancutting,zhong2026achieving}. In addition to traditional zero-shot prompting, instructional prompting has been shown to further enhance LLMs' reasoning capabilities for these problems \cite{xucan}. However, LLMs perform suboptimally on complex mathematical reasoning, often struggling with problems that require multi-step calculations, handling irrelevant information, or applying real-world knowledge not explicitly stated in the question \cite{srivatsa2024makes}.

\begin{figure*}[t]
\centering
\includegraphics[width=\textwidth]{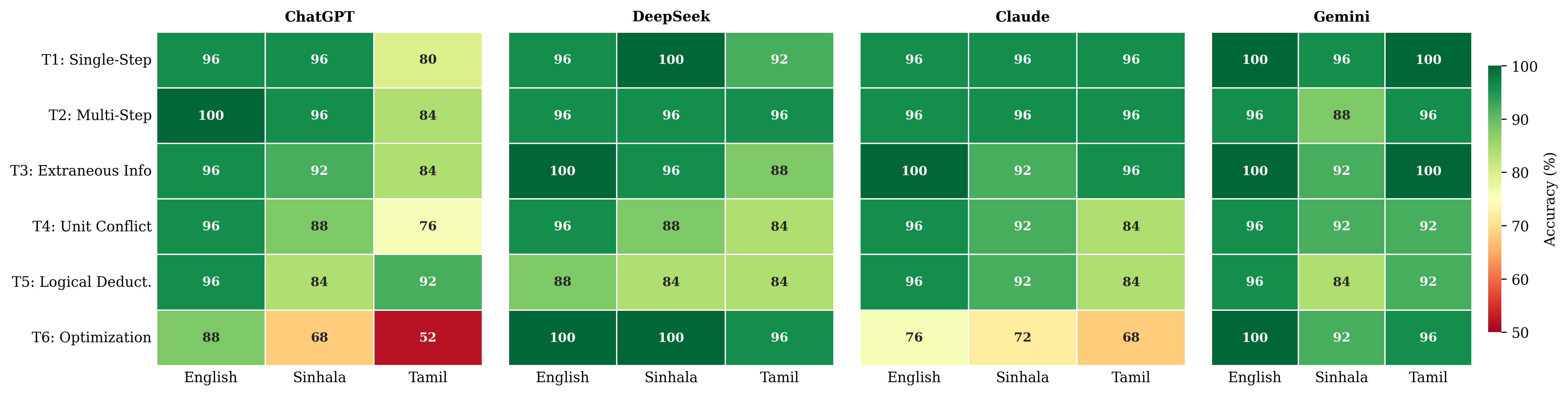}
\caption{Accuracy (\%) across four LLMs, six problem types, and three languages. Darker green indicates higher accuracy; red tones highlight degradation.}
\label{fig:heatmap}
\end{figure*}

\subsection{Multilingual Mathematical Evaluation}

While influential works on LLMs and mathematical reasoning mainly focus on English, the most prominent multilingual benchmark, MGSM \cite{shi2023language}, evaluates LLMs across ten languages using translated GSM8K problems \cite{cobbe2021training}. However, its reliance on translation introduces artifacts and its single-score evaluation conflates distinct reasoning skills. Recent investigations into multilingual LLM behavior suggest that models may not genuinely reason in each target language. Zhao et al.\cite{zhao2024llama} found evidence that LLMs route multilingual inputs through English-centric processing pathways, while Etxaniz et al.\cite{etxaniz2024bertaqa} demonstrated performance asymmetries consistent with internal translation to English representations. These findings raise questions about whether apparent multilingual mathematical competence reflects genuine cross-lingual reasoning or dependence on English-mediated processing.

\subsection{Sinhala, and Tamil Evaluation}

Evaluation of LLMs in South Asian languages, particularly Sinhala and Tamil, remains limited. Jayakody et al.\cite{jayakody2024performance} explored LLM performance on Sinhala, finding that while some models demonstrate strong multilingual capabilities, others exhibit poor performance but remain amenable to improvement through fine-tuning. Pramodya et al.\cite{pramodya2025sinhalammlu} introduced a Sinhala benchmark, finding limited performance particularly in culturally rich domains, and demonstrated that translated benchmarks fail to capture domain-specific terminology and cultural context. Mathematical reasoning evaluation in Tamil remains largely unexplored, with existing work focusing on general language tasks \cite{balachandran2023tamil,ponnusamy2023tamil}.

\subsection{LLMs as Educational Tools}

AI tutoring systems powered by LLMs are increasingly adopted in schools and universities. Platforms such as Khan Academy's AI assistant use LLMs to guide students through problems step by step \cite{fuligni2025would}, and similar tools are being trialled for homework support and automated assessment in many countries \cite{kasneci2023chatgpt,wardat2023chatgpt,alhatmi2024exploring}. Many of these AI tutoring systems have been developed primarily in English, raising concerns about their effectiveness in multilingual educational settings, especially for low-resource languages such as Tamil and Sinhala. Students in these language contexts may receive incorrect guidance from tools that are not evaluated for their specific needs. This situation is further complicated by the limited support of digital resources in these communities. Therefore, using AI tutoring tools without language-specific evaluation could disadvantage the very students they aim to help. Currently, there has been very limited assessment of large language models for mathematical reasoning in Sinhala or Tamil within education, highlighting a critical gap in understanding their effectiveness for these languages.

Our work addresses these limitations by providing the first systematic evaluation of mathematical reasoning across multiple problem types in both Sinhala and Tamil, while enabling fine-grained analysis that can provide behavioral evidence for language-dependent processing patterns in mathematical reasoning.

\section{Methodology}\label{sec:methodology}

We present our framework for assessing multilingual mathematical reasoning in LLMs, covering our problem taxonomy, dataset, and evaluation protocol.

\subsection{Problem Taxonomy}
\label{sec:taxonomy}

We represent each math word problem as $P = (Q, N, R, A)$: $Q$ is the question text, $N = \{n_1, n_2, \ldots, n_k\}$ is the set of numerical values, $R$ is the set of required operations, and $A$ is the correct answer. Our six types are defined based on the structure of $R$ and how $N$ relates to the solution.

\subsubsection{Type 1: Single-Step Problems}
$|R| = 1$: one arithmetic operation on numbers from $N$ gives $A$. This serves as a baseline where models that struggle here likely have fundamental issues.

\subsubsection{Type 2: Multi-Step Problems}
$|R| > 1$ with sequential operations where $r_i \in R$ feeds into $r_{i+1}$. These require tracking intermediate results across a chain of calculations, making this useful for testing sustained numerical reasoning.

\subsubsection{Type 3: Problems with Extraneous Information}
The problem includes irrelevant numbers: $N_r \subset N$ contains the relevant values and $N_d = N \setminus N_r \neq \emptyset$ the distractors. Models must identify which numbers matter before calculating. This is particularly interesting in Sinhala and Tamil, where complex morphology and flexible word order can make signal-noise separation harder.

\subsubsection{Type 4: Unit Conflict Problems}
Some relevant quantities have mismatched units: there exist $n_i, n_j \in N_r$ with $unit(n_i) \neq unit(n_j)$ that must be used together, requiring conversion first. We test whether the problem's language affects how reliably models handle such conversions.

\subsubsection{Type 5: Logical Deduction Problems}
These questions can't be solved through direct arithmetic. The text describes relationships between unknowns, and models must build equations $f(x_1, x_2, \ldots, x_m) = 0$ from the verbal descriptions in $Q$ and solve them. The main challenge is converting natural language into algebra, which can vary based on how languages express quantitative relationships.

\subsubsection{Type 6: Optimization Problems}
In these questions, models must find the maximum or minimum of a function over constraints defined in the problem text. This requires extracting objectives and constraints from prose, formulating the problem mathematically, and applying optimization techniques. We expect the largest cross-lingual gaps here due to the complexity. 

Note that within each type, problems were selected to reflect diverse contexts and surface forms, ensuring broad coverage of the reasoning competency targeted by that type. Table~\ref{tab:taxonomy} summarizes the six question types, and Figure~\ref{fig:questions} illustrates examples from each across the three languages.

\begin{figure*}[t]
\centering
\includegraphics[width=\textwidth]{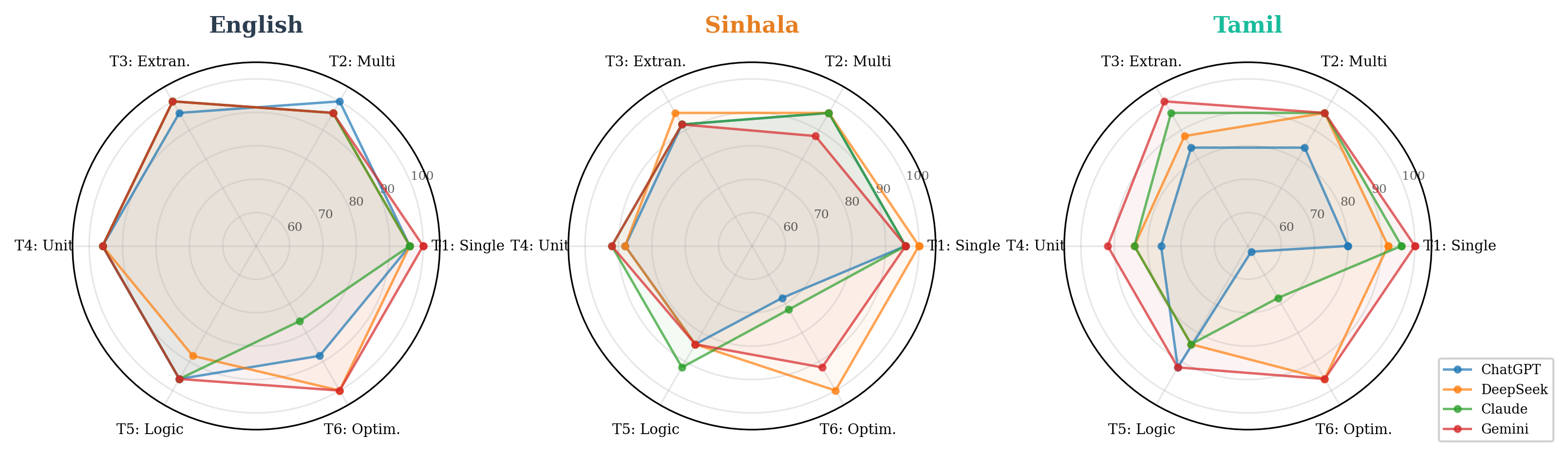}
\caption{Radar plots comparing model accuracy (\%) across six problem types for each language. Polygon shrinkage from English to Sinhala and Tamil reflects cross-lingual performance loss.}
\label{fig:radar}
\end{figure*}

\subsection{Alignment with Educational Levels}
\label{sec:taxonomy_education}

The proposed taxonomy corresponds to the different stages of mathematical learning in education. Type 1 (Single-Step) and Type 2 (Multi-Step) problems are primarily found in primary and early secondary education, where students learn basic arithmetic and procedural skills. Type 3 (Extraneous Information) and Type 4 (Unit Conflict) are usually encountered in middle school, requiring students to navigate word problems, filter out unnecessary information, and manage unit conversions, which are the skills that blend math with language understanding. Type 5 (Logical Deduction) reflects the algebraic thinking developed in secondary school, where students translate descriptions into equations and find unknowns. Lastly, Type 6 (Optimization) relates to advanced secondary or early tertiary-level problem-solving, focusing on constrained reasoning and mathematical modeling. This taxonomy evaluates not just model performance but also the cognitive challenges faced by students at various stages, making it effective for assessing the educational reliability of models in learning contexts.

\begin{table}[t]
\centering
\caption{Summary of the six math word problem types and targeted reasoning competencies.}
\label{tab:taxonomy}
\resizebox{0.9\linewidth}{!}{%
\begin{tabular}{cll}
\toprule
\textbf{Type} & \textbf{Category} & \textbf{Reasoning Competency} \\
\midrule
1 & Single-Step & Basic arithmetic comprehension \\
2 & Multi-Step & Chained reasoning \\
3 & Extraneous Information & Distractor filtering \\
4 & Unit Conflict & Unit conversion reasoning \\
5 & Logical Deduction & Algebraic formulation \\
6 & Optimization & Constrained optimization \\
\bottomrule
\end{tabular}%
}
\end{table}

\subsection{Dataset Construction}
\label{sec:dataset}
Rather than translating existing benchmarks, we write each problem natively in all three languages. This avoids translation artifacts that mix language ability with translation quality~\cite{shi2023language}. Semantic equivalence across language versions is properly verified, without requiring word-for-word correspondence. This approach also preserves naturally occurring language use, ensuring that problems reflect how students actually encounter mathematics in their native language contexts. The final dataset has 25 problems per type per language, totaling $6 \times 25 \times 3 = 450$ instances.

\subsection{Evaluation Protocol}
\label{sec:protocol}

We test four model configurations from different providers using zero-shot prompting: GPT-4o (OpenAI) \cite{islam2025gpt}, DeepSeek-V3 (DeepSeek-AI) \cite{zhao2025insights}, Gemini~2.5 (Google) \cite{comanici2025gemini}, and Claude Sonnet~4 (Anthropic) \cite{anthropic2023introducing,anthropic2025claudesonnet4}. We score outputs by extracting the final numerical answer and comparing it to the ground truth $A$. 

Zero-shot prompting \cite{li2023practical} reflects typical student usage, where queries are posed without prompt engineering. We adopt this setting to evaluate the models under realistic and minimally assisted conditions, providing a more faithful estimate of their reliability in educational use. We further analyse performance across problem types and languages to identify systematic patterns of cross-lingual degradation.

\section{Results and Discussion}
\label{sec:results}

Figure~\ref{fig:heatmap} shows each model's accuracy across all six problem types and three languages. Figure~\ref{fig:radar} illustrates how each model's reasoning profile changes across languages.

\subsection{Overall Cross-Lingual Performance}
\label{sec:overall}

All four models achieve higher accuracy in English. However, performance drops in Sinhala, with most models showing even larger declines in Tamil (Figure~\ref{fig:heatmap}). ChatGPT shows the steepest drop, falling from 95.3\% in English to 78.0\% in Tamil. DeepSeek proves most stable with 94\%, and 90\% accuracy in Sinhala and Tamil, respectively. Interestingly, Gemini performs better in Tamil (96.0\%) than Sinhala (90.7\%), possibly reflecting differences in training data composition or script processing.

\subsection{Problem Type Analysis}
\label{sec:type_analysis}

The degradation varies significantly across problem types, a key finding highlighted by the taxonomy.

\subsubsection{Robust categories} Types 1-2 (Single-Step, Multi-Step) remain strong across languages. Claude and DeepSeek maintain near-perfect scores in all three languages, showing that basic arithmetic reasoning transfers well regardless of surface language.

\subsubsection{Consistently vulnerable category} Unit conflict problems (Type 4) show the most consistent cross-lingual drops across all models. ChatGPT has the largest gap at 20\%, but all models lose accuracy, suggesting that interpreting and converting units across different scripts and morphological systems poses a systematic challenge.

\subsubsection{Optimization as a stress test} Type 6 reveals the starkest differences. ChatGPT's performance plummets from 88\% to 52\% in Tamil, the largest drop in our evaluation. Claude also declines from 76\% to 68\% in Tamil. In contrast, DeepSeek performs nearly perfectly across all languages, and Gemini also holds steady. The radar plots in Figure~\ref{fig:radar} clearly show how ChatGPT's performance profile distorts when moving from English to Tamil, while other models maintain their shape.

\subsection{Qualitative Analysis of Failures}

\begin{figure}
\centering

\begin{subfigure}{\linewidth}
    \centering
    \includegraphics[width=\linewidth]{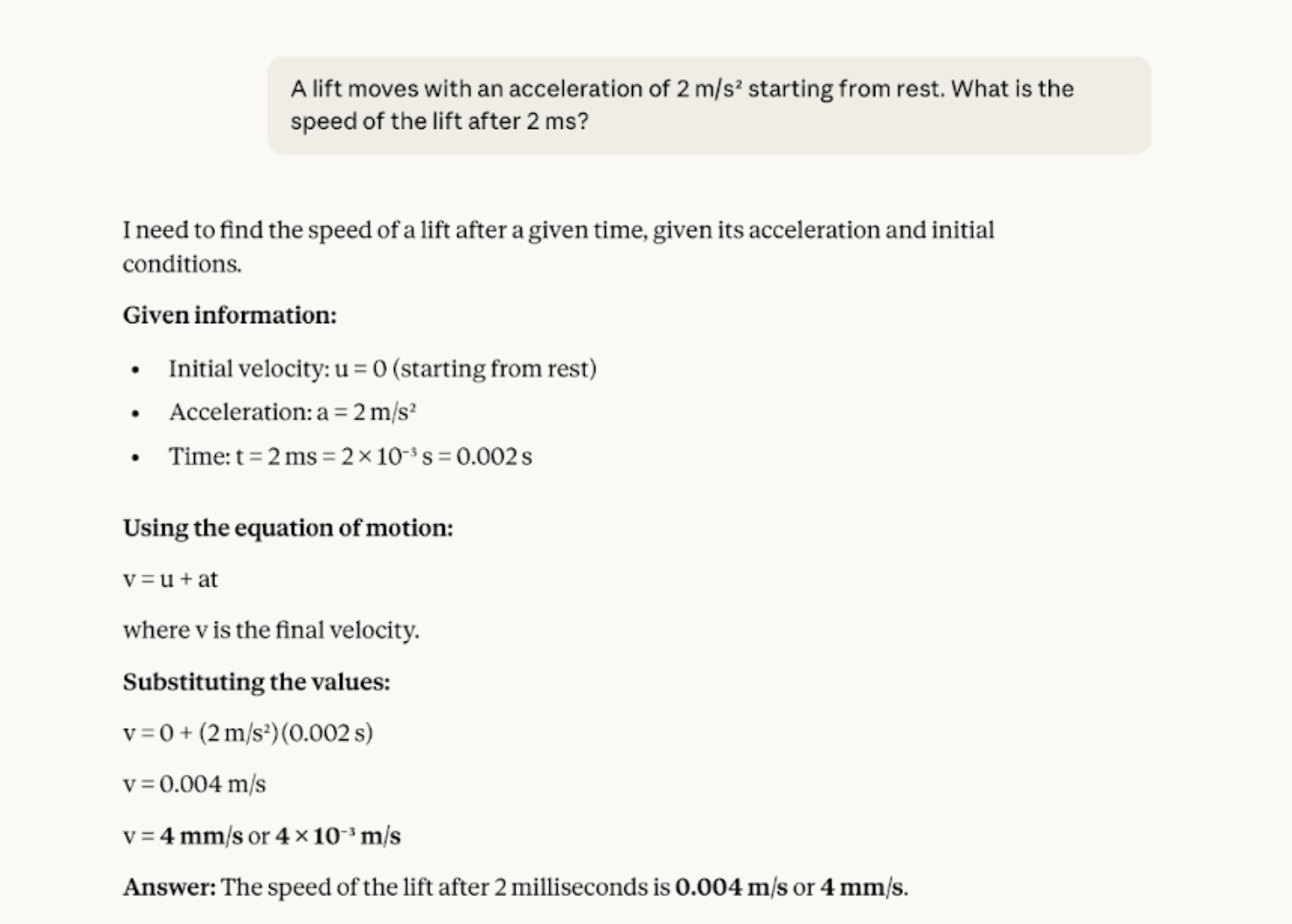}
    \caption{English}
\end{subfigure}

\vspace{0.4em}

\begin{subfigure}{\linewidth}
    \centering
    \includegraphics[width=\linewidth]{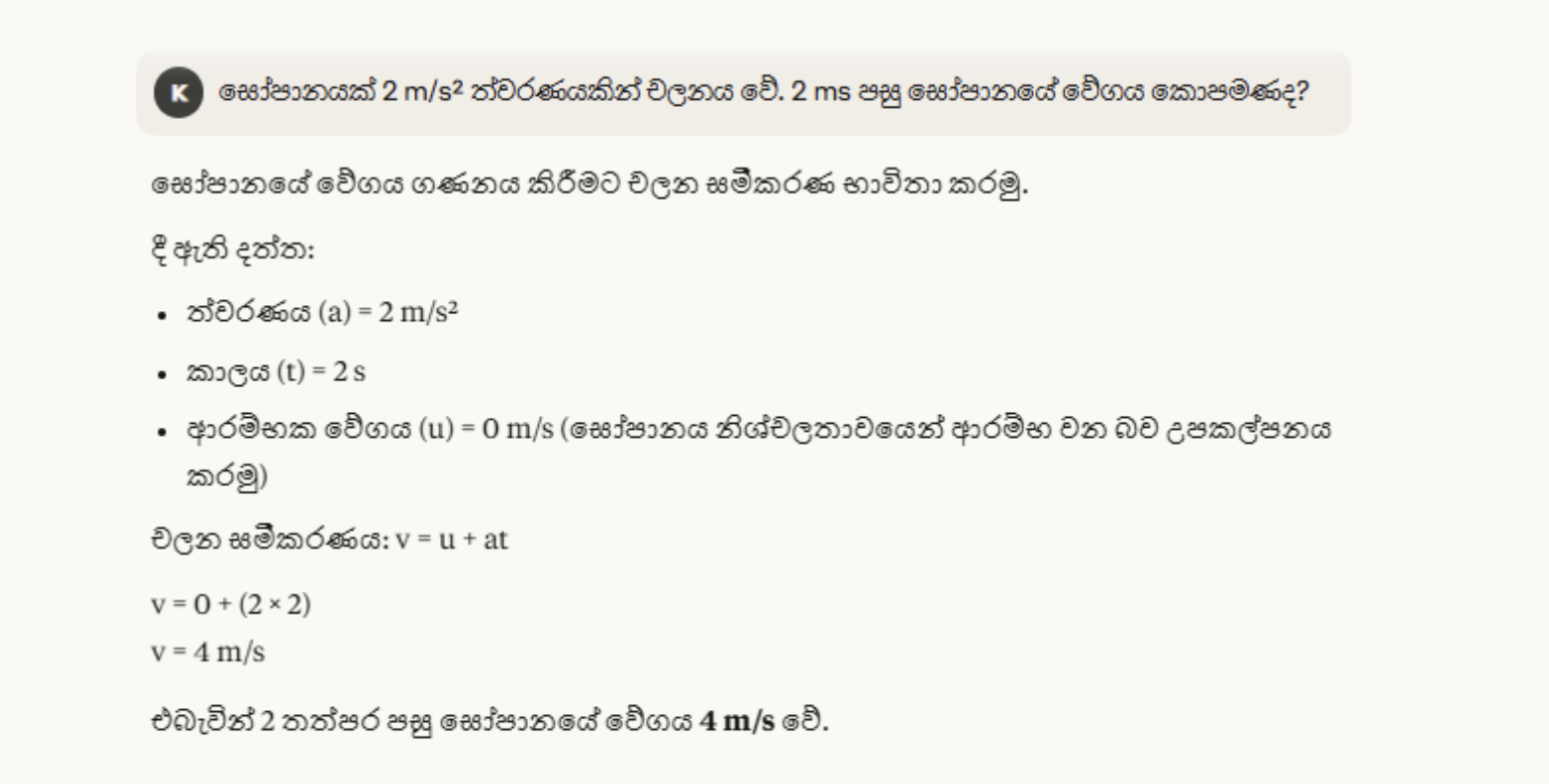}
    \caption{Sinhala}
\end{subfigure}

\vspace{0.4em}

\begin{subfigure}{\linewidth}
    \centering
    \includegraphics[width=\linewidth]{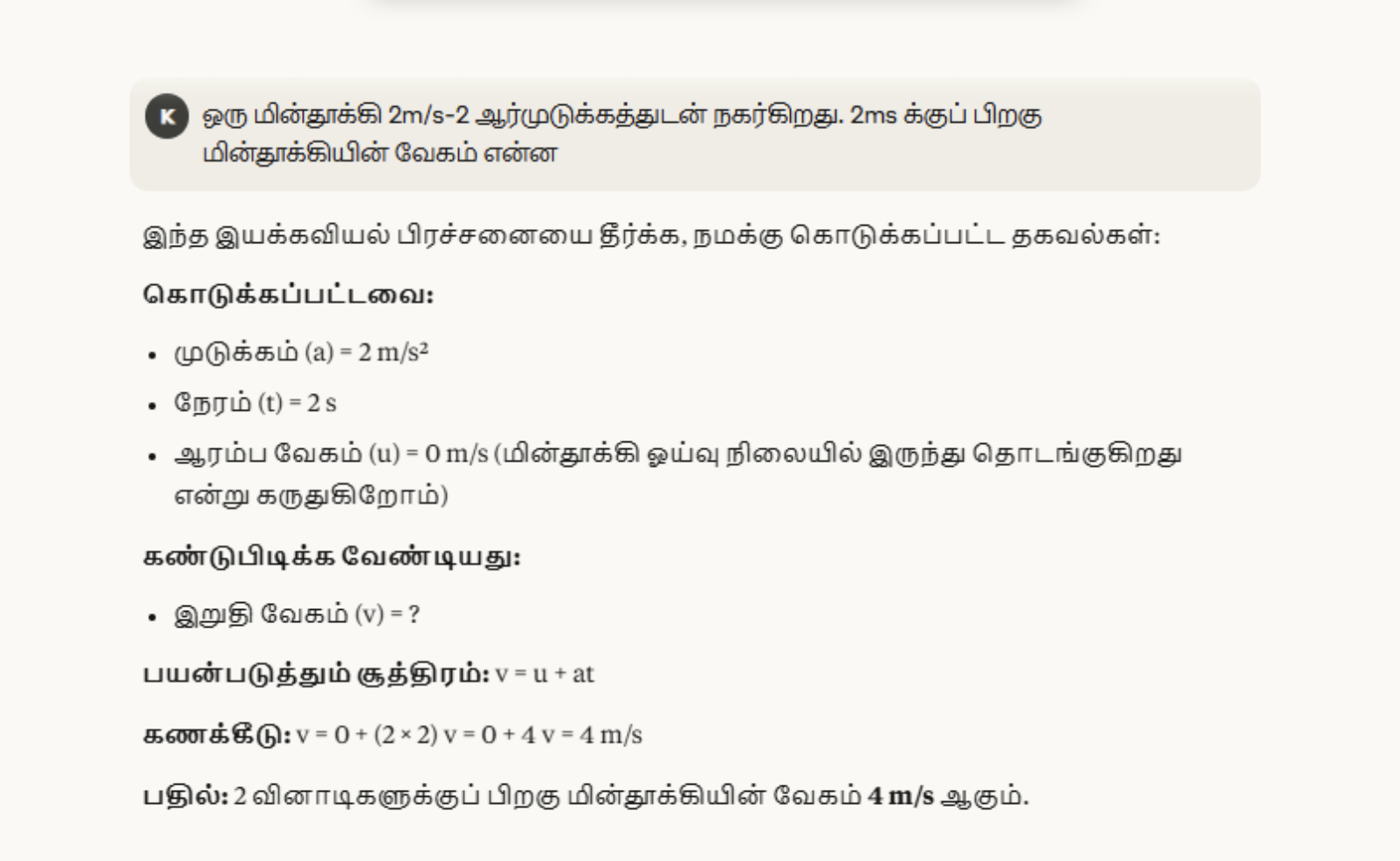}
    \caption{Tamil}
\end{subfigure}

\caption{Representative example of a unit conflict problem (Type 4) posed to the same LLM (Claude) across three languages. In English, the model correctly identifies that ``2 ms'' denotes milliseconds and converts accordingly (v = 0.004 m/s). In Sinhala, the model ignores the unit conversion and treats the time as 2 seconds, yielding the incorrect answer of 4 m/s. Tamil produces the same error. This pattern, where reasoning is correct in English, but unit misinterpretation occurs in low-resource languages, illustrates that cross-lingual degradation manifests not as arithmetic failure but as a failure to parse linguistically embedded unit information. Screenshots reproduced for research purposes.}
\label{fig:qualitative}
\end{figure}

To complement the quantitative results, we include an example illustrating common LLM failure modes in multilingual mathematical reasoning. We have selected a unit conflict problem that is correctly solved in English, but incorrectly solved in both Sinhala and Tamil, as shown in Figure \ref{fig:qualitative}.

 This example demonstrates that while models typically perform well in English, their responses in low-resource languages show incorrect reasoning. These instances reflect broader patterns observed in our evaluation, rather than being isolated cases.

\section{Conclusion, Limitations, and Future Work}
\label{sec:conclusion}

Our analysis reveals that large language model performance in Sinhala and Tamil depends critically on the problem type. For basic arithmetic, models achieve nearly identical accuracy across languages. However, for complex types like unit conversion and optimization, performance drops sharply, raising questions about how reliably these models handle non-English mathematical reasoning. The observed pattern, where simple computation transfers but linguistically demanding reasoning fails, may reflect several factors: differences in training data coverage across mathematical registers, tokenisation challenges, or language-specific processing behaviours. Disentangling these factors remains an important direction for future work.

From an educational standpoint, our findings carry an important practical message. A model that scores 88\% accuracy in English may drop to 52\% on the same problem type in Tamil. Schools and education technology developers deploying large language models as tutoring or assessment tools in Tamil- or Sinhala-medium settings should carry out language-specific testing, especially for problem types involving unit conversion and multi-step reasoning, which are central to standard school curricula. Our taxonomy provides a starting point for such evaluation. Furthermore, based on the accuracy, educational staff such as teachers should not rely solely on the outcome of LLMs, since they are not entirely accurate. For instance, accuracy as low as 52\% was observed for optimization problems in Tamil, which is insufficient for reliable student-facing use. Therefore, teachers should be aware that there is a risk of giving the wrong answer and should independently verify answers for these problem types. They can also try few-shot prompting or chain-of-thought prompting to enhance accuracy.

Our work has several limitations. The dataset of 25 problems per type, while adequate for revealing broad patterns, limits the detection of small differences between models or languages. We test only zero-shot prompting; few-shot or chain-of-thought approaches might change these patterns. We also evaluate models as black boxes without access to internal representations, so the translation-dependent processing hypothesis remains an inference from behavioral evidence rather than a mechanistic finding. In future work, we plan to expand our dataset to cover a broader range of school curriculum topics. Also, we plan to use interpretability methods to probe whether models rely on translation-based processing during mathematical reasoning.


\section*{Acknowledgement}

The authors used AI-based tools solely for language polishing 
and writing clarity. All AI-assisted edits were reviewed and 
approved by the authors to ensure accuracy and integrity of 
the content. The research design, dataset construction, 
experimental analysis, and all conclusions were conducted and 
verified by the authors.



\bibliography{references} 
\bibliographystyle{ieeetr}

\end{document}